\documentclass[twocolumn,letterpaper]{IEEEAerospaceCLS}  

\usepackage{graphicx}   
\usepackage{amsmath}
\usepackage{amsfonts}
\usepackage{enumitem}

\usepackage[font=footnotesize,justification=justified,singlelinecheck=false]{caption}
\begin{document}
\title{Terrain-aware Low Altitude Path Planning}

\author{%
Yixuan Jia \\
Massachusetts Institute of Technology \\
 77 Massachusetts Ave \\
Cambridge, MA 02139\\
yixuany@mit.edu
\and 
Andrea Tagliabue\\
Massachusetts Institute of Technology \\
 77 Massachusetts Ave \\
Cambridge, MA 02139\\
atagliab@mit.edu \\
\and
Annika Thomas\\
Massachusetts Institute of Technology \\
 77 Massachusetts Ave \\
Cambridge, MA 02139\\
annikat@mit.edu
\and
Navid Dadkhah Tehrani\\
Lockheed Martin Corporation | Sikorsky Aircraft \\
6900 Main Street \\
Stratford, CT 06615\\
navid.dadkhah.tehrani@lmco.com \\
\and
Jonathan P. How\\
Massachusetts Institute of Technology \\
 77 Massachusetts Ave \\
Cambridge, MA 02139\\
jhow@mit.edu \\
\thanks{}              
}

\maketitle

\thispagestyle{plain}
\pagestyle{plain}

\maketitle

\thispagestyle{plain}
\pagestyle{plain}

\begin{abstract}
    In this paper, we study the problem of generating low-altitude path plans for nap-of-the-earth (NOE) flight in real time with only RGB images from onboard cameras and the vehicle pose. 
    We propose a novel training method that combines behavior cloning and self-supervised learning, where the self-supervision component allows the learned policy to refine the paths generated by the expert planner.
    Simulation studies show $24.7 \%$ reduction in average path elevation compared to the standard behavior cloning approach.
\end{abstract}

\tableofcontents

\section{Introduction}
Nap-of-the-earth (NOE) flight is an important tactic to reduce the exposure of an aircraft during flights. For a piloted aircraft flying at high speed, NOE flights are intensive as they require the pilots to extract terrain information and react very quickly to new information. Therefore, it would be beneficial to automate some of the tasks, such as path planning \cite{cheng1988considerations}. Moreover, solutions relying only on sensors that do not emit (e.g. cameras) are preferred since they further reduce the exposure of the aircraft.

Automating part of the navigation task for NOE flights has been studied in numerous previous works. Cheng and Sridhar \cite{cheng1988considerations} proposed a guidance system architecture for NOE flights which considers the functionality one would need for different levels of the task, from waypoint selection to path generation and finally to control generation. 
Menon et al. \cite{menon1991optimal} proposed a trajectory planning method for terrain-following flight, which is a simplified problem of NOE flight.
Similarly, Lu and Pierson \cite{lu1995optimal} studied off-line trajectory generation for terrain-following flight where the nonlinear dynamics were taken into account.
Trajectory tracking for NOE flight was studied in \cite{lapp2004model,hess1989application,lu1995aircraft} using predictive control. 
To the best of our knowledge, most previous works either focused on off-line planning or relied on sensors that emit, such as LiDAR~\cite{johnson2014comparison}.

\begin{figure}
\centering
\includegraphics[trim={6cm 0cm 6cm 0cm},clip,width=8.5cm]{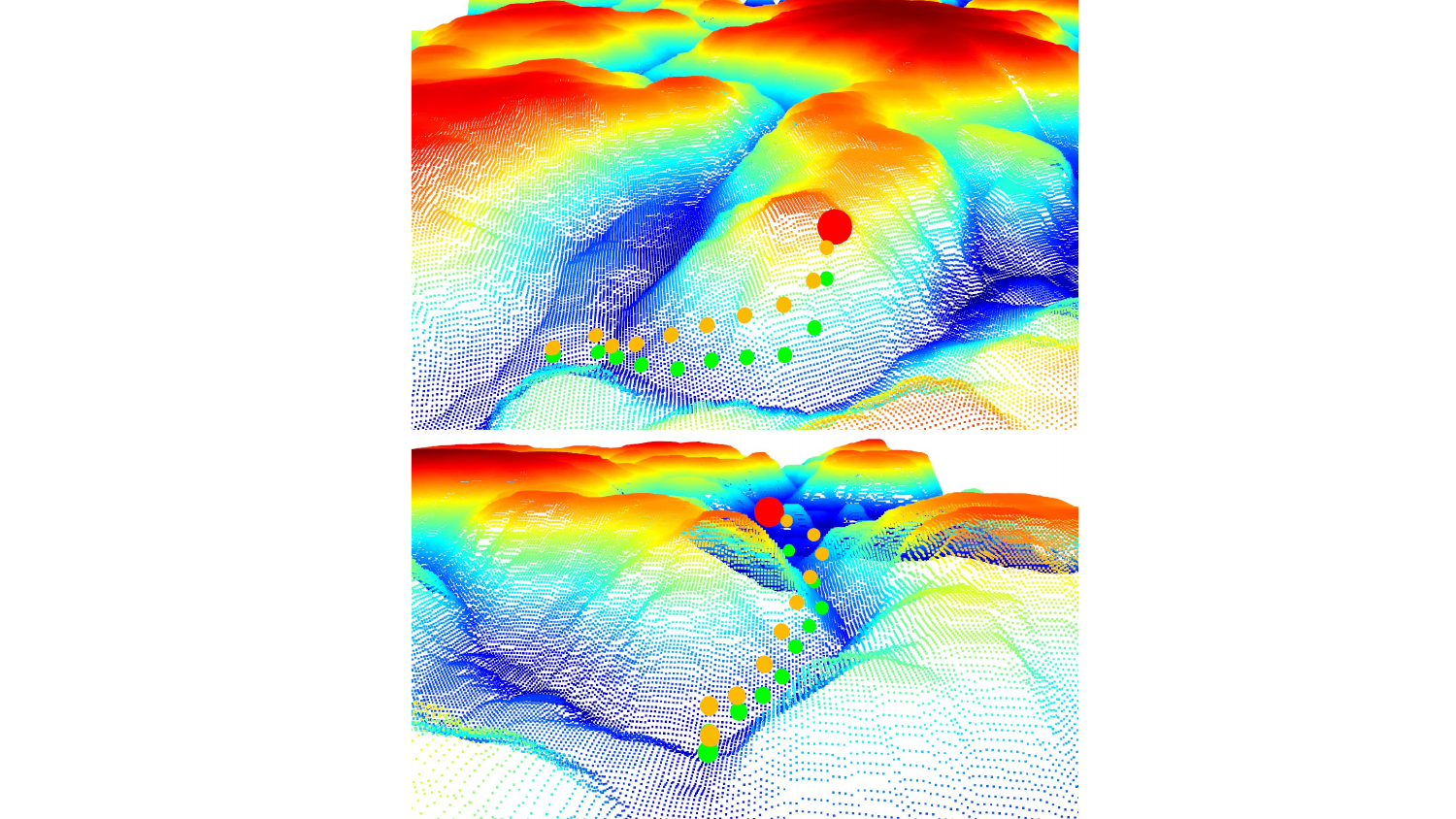}
\caption{Example of the expert path (orange) vs. the path generated by our policy (green). The red dot denotes the goal. Note that our policy is able to generate a path lower than the path from the expert planner.}
\label{fig:expert_vs_student}
\end{figure}

\begin{figure*}[t]
\centering
\includegraphics[trim={0cm 3cm 0cm 1.5cm},clip,width=16cm]{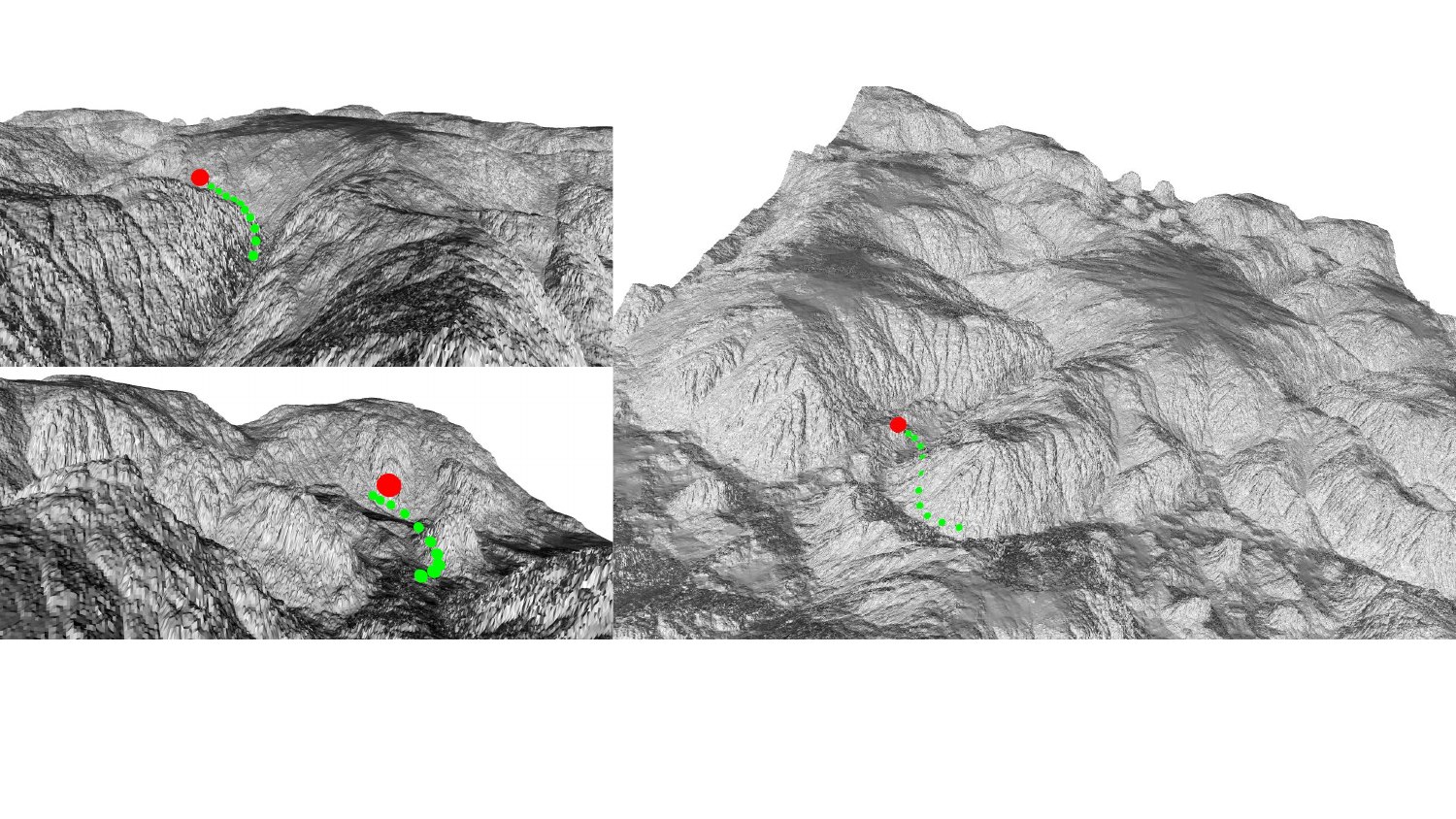}
\caption{Example of paths generated by our policy, marked in green color. The red dots denote the goals. Note that instead of climbing over the terrain to reach the goal, which would result in paths with possibly shorter length but also higher elevation, the policy is able to exploit the terrain and plan paths with lower elevations.}
\label{fig:student_paths}
\end{figure*}

In recent years, the advancement of deep learning has unlocked the potential of light-weight onboard computation for vision-based planning and control tasks \cite{shah2022viking,tagliabue2024efficient,loquercio2021learning}. Two of the most common approaches are imitation learning and reinforcement learning, where the former relies on an expert planner providing demonstrations while the latter learns by trial and error. Imitation learning is known to be easier to train, but the performance relies heavily on the expert planner since it essentially learns to copy the expert's behavior without explicit knowledge of the quality of demonstrations provided by the expert \cite{cheng2018fast,sun2018truncated}. On the other hand, reinforcement learning can theoretically learn an optimal policy given enough exploration, but can be extremely data-hungry in practice, and the reward design and hyperparameter tuning can be difficult \cite{ibarz2021train}. 
Recently, another approach using self-supervised learning was proposed that essentially embeds the optimization involved in planning and control directly into the training, in the hope of achieving better data efficiency and policy performance \cite{yang2023iplanner,roth2024viplanner}, thus potentially outperforming both imitation learning and reinforcement learning. However, as will be discussed in this paper, this approach requires extra regularizations (compared to imitation learning) to produce desirable results.




In this paper, we consider the task of online local path planning using only onboard RGB cameras and aircraft poses.
Our approach is inspired by self-supervised learning methods but combines the strength of both imitation learning and self-supervised learning for NOE flights. Our contributions include:
\begin{itemize}[left=1em, topsep=0pt, partopsep=0pt, parsep=0pt, itemsep=0pt]
    \item A training pipeline that learns a policy capable of generating low-altitude path plans with only RGB images and aircraft poses.
    \item A novel supervision method that is capable of generating policies that outperform student policies trained with standard behavior cloning approach on our task.
\end{itemize}
We will present our methodology in section 2, where the design of the expert planner and the core of our training method will be introduced. The details of practical implementation will be presented in section 3, which covers dataset preparation, student policy architecture, and the training objective design. Finally, simulation studies will be presented in section 4.



\section{Methodology}

We present our method in this section. Recall that the objective of NOE flights is to plan a path that reaches the goal from the current location while maintaining low altitude to avoid exposure. Our pipeline has two major components: an expert planner that is used to generate demonstrations of path plans and a learning module that distills and refines the expert planner. We start by setting up notation and some definitions.

\subsection{Notations}
We use ${\mathbb{R}}$ to denote real numbers and ${\mathbb{S}}^1$ to denote the unit circle (used to represent rotation angles). We also use $[n] := \{0, 1, \dots, n\}$.

\subsection{Terrain Representation}
In some parts of our proposed method, different representations of the terrain are used for different purposes. 
We set up and clarify the notation in this subsection.
Given a terrain represented as a point cloud ${\mathcal{T}} = \{(x_1, y_1, z_1), (x_2, y_2, z_2), \dots, (x_{|\mathcal{T}|}, y_{|\mathcal{T}|}, z_{|\mathcal{T}|})\}$, we can convert it to an elevation map \cite{fankhauser2016universal}, which is essentially a function $\texttt{Elevation}_{{\mathcal{T}}}: {\mathbb{R}}^2 \to {\mathbb{R}}$ where $\texttt{Elevation}_{{\mathcal{T}}}(x, y)$ represents the elevation of the terrain at $(x, y)$.

\subsection{Expert Planner Design}
Our expert planner is based on \cite{lim2024safe}, which is a  sampling-based planner that uses the Dubins airplane model \cite{chitsaz2007time}. The planner takes two layers of grid maps, $\texttt{Elevation}_{\min}$ and $\texttt{Elevation}_{\max}$, \cite{fankhauser2016universal,triebel2006multi} that constrain the minimum and maximum elevation of the vehicle with respect to the terrain. In other words, for each position $(x, y, z)$ on the path, the constraint $\texttt{Elevation}_{\min}(x, y) \le  z \le \texttt{Elevation}_{\max}(x, y)$ must be satisfied.
Then, given a pair of start and goal locations, the planner computes a safe path between the start and the goal assuming constant speed. 
The safe path is computed with a variant of RRT* \cite{karaman2011sampling}. 
Since the original work focuses on path planning for safe flights, in order to make it applicable to NOE flights, we modify the objective of the planner to be a weighted combination of the altitude of the path and the length of the path, to encourage the planner to generate low-altitude paths while avoiding making long detours.
To be more specific, suppose the path is given by $p(\cdot) : [T] \to {\mathbb{R}}^{3} \times {\mathbb{S}}^1$, where ${\mathbb{R}}^{3} \times {\mathbb{S}}^1$ (position and yaw) represents the state space of Dubins airplane, and $T$ denotes the total number of time stamps of the path. Then the cost of the path is:
\begin{align}\label{eq:expert_cost}
    {\mathcal{L}}(p(\cdot)) = \alpha \cdot l(p(\cdot)) + \beta \cdot \sum_{t=0}^T p_z(t)
\end{align}
where $l(\cdot)$ computes the length of the path, $p_z(\cdot)$ denotes the z-coordinate of the path, and $\alpha, \beta > 0$ are hyperparameters that balance the two terms.


\subsection{Combining Behavior Cloning and Self-supervised Learning}
In our early experiments, we found that the expert was able to find feasible solutions in a reasonable amount of time (within $45$s for our problem) but struggled to converge to optimal or near-optimal solutions even given $600$s. The authors of \cite{lim2024safe} observed similar issues. 
Moreover, since in our case low-altitude paths are preferred, we tried to set both 
the minimum and maximum allowed elevations to be low to force the expert planner to find paths with low average altitude. But we found that it made the expert struggle to find feasible paths due to a combination of steep terrain and constraints on the turning radius, climbing rate, and altitude. However, relaxing the altitude constraint by setting the maximum elevation to be a higher value helped mitigate this issue. We observed that the resulting expert paths contain positions with reasonable $x, y$ coordinates (i.e. $\texttt{Elevation}_{{\mathcal{T}}}(x, y)$ is relatively low), but $z$ is usually conservative(i.e. $z > \texttt{Elevation}_{{\mathcal{T}}}(x, y)$ by a considerable amount).
If we train the student policy using the standard BC approach (i.e. directly copying the expert), then the student policy won't be able to generate low-altitude paths, since the expert demonstrations do not contain enough demonstrations of low-altitude paths. 
Instead, we propose a supervision method that combines BC and self-supervision.

We use BC to supervise the generated $x, y$ coordinates (i.e. planar motion) by minimizing deviations from the $x, y$ coordinates generated by the expert. To account for possible multi-modalities of expert demonstrations, the relaxed winner-takes-all loss is  used to assign loss \cite{loquercio2021learning}.

To achieve lower altitude flight by the student policy, we use self-supervision to directly supervise the planned $z$-coordinates via a loss term:
\begin{align}\label{eq:ss_loss}
    \|z - (\delta z_{\text{des}} + \texttt{Elevation}_{{\mathcal{T}}}(x, y))\|_2^2
\end{align}
where $x, y, z$ denotes the planned position of the student policy and $\delta z_{\text{des}}$ denotes the desired relative elevation to the terrain, which we set to be a constant. This is computed for every point on the planned path for all the planned paths, but we write the expression for only one point to simplify the notation.

{\bf Remarks}
A different approach would be to consider training a policy in a fully self-supervised fashion by converting the elevation map to a signed distance field (SDF), similar to \cite{yang2023iplanner}. One would expect that the optimal planar motion (i.e., $x, y$ coordinates) could be obtained using the gradient information provided by the SDF. While possible, we find such an approach introduces extra challenges during training (e.g. extra loss terms need to be added to regularize/shape the output of the policy). 
For our problem, we found the need to add at least the following loss terms:
\begin{itemize}[left=1em, topsep=0pt, partopsep=0pt, parsep=0pt, itemsep=0pt]
    \item Term that penalizes the deviation of the last predicted waypoint from the goal to encourage the policy to reach the goal rather than simply generating $(x, y)$'s where $\text{Elevation}_{\mathcal{T}}(x, y)$ are low
    \item Term that penalizes infeasible plans (e.g. waypoints spread all over low-elevation areas and far from each other)
    \item Term that penalizes degenerate plans (e.g. the last waypoint reaches the goal while all other generated waypoints are at the same location $(x, y)$ where $\texttt{Elevation}_{\mathcal{T}}(x, y)$ is low).
\end{itemize}
Therefore, we would end up having at least three more loss terms (and the weights for each of them) just to regularize the output of the policy. Additionally, there does not appear to be a clear and intuitive strategy on how to tune the weights.
On the other hand, providing planar motions from the expert removes the need for those terms and greatly simplifies the training process. 
The BC part essentially identifies modes of the distribution (of optimal paths) that we want the student policy to learn from, and the self-supervision part refines within the modes, which is what gradient-based methods like deep learning are good at.

\section{Implementation}

\subsection{Dataset Preparation}
Suppose we have collected $N$ paths generated by the expert represented as $$\{(x^n_t, y^n_t, z^n_t, {q}^n_{x, t}, {q}^n_{y, t}, {q}^n_{z, t}, {q}^n_{w, t})_{t \in [T_n]}\}_{n \in \{1, \dots, N\}},$$ where $(x^n_t, y^n_t, z^n_t, {q}^n_{x, t}, {q}^n_{y, t}, {q}^n_{z, t}, {q}^n_{w, t})$ represents the pose (position and orientation) of the vehicle at time stamp $t$ of the $n$-th collected path. 

\begin{figure}[t]
\centering
\includegraphics[trim={7cm 10cm 5cm 10cm},clip,width=8cm]{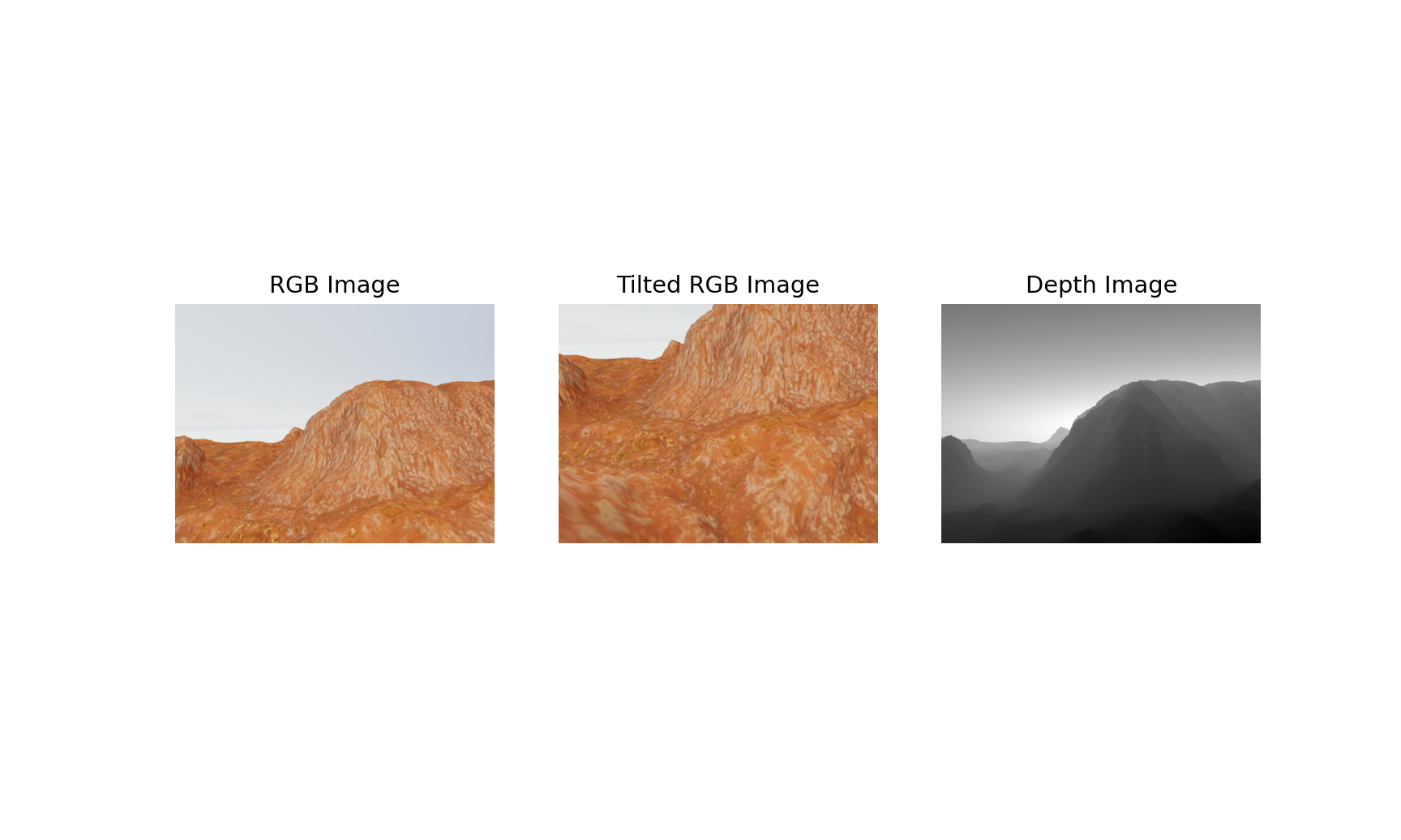}\\
\caption{{Example of collected images. At each planned vehicle pose, we collect two RGB images and a depth image. We collect a RGB image (left) and its corresponding depth image (right) from a forward facing camera, and another RGB image (middle) from a downward facing camera to capture extra information.}}
\label{fig:img_data}
\end{figure}

We can then use a simulator that supports camera image rendering to collect camera images that will be used for training. In our case, we use NVIDIA\textsuperscript{\tiny\textregistered}
 Isaac Sim \cite{isaacsim} to render camera images. For each collected pose, we place a forward-facing camera at that pose to collect an RGB image and a depth image. We additionally set a camera that is tilted downward by $30 \deg$ (with respect to the forward-facing camera) and use it to collect downward-facing RGB images to provide extra information. Therefore, for each pose in the collected path, we collect three images: two RGB images and a depth image. An example of collected image data is shown in Fig \ref{fig:img_data}. To help the policy focus on things that are closer, we take the $\log$ of the depth image, inspired by \cite{he2024agile}. 
 
To facilitate training, we interpolate the collected paths and subsample the states at a fixed distance $d$. In practice, we find that doing so helps the training converge faster and also makes the training more stable compared to the case where we do not perform interpolation.
For each sampled state $s_t = (x_t, y_t, z_t, {q}_{x, t}, {q}_{y, t}, {q}_{z, t}, {q}_{w, t})$, we compute a heading vector $h_t$ by taking the difference between the position that is $H$ steps ahead in the path, $(x_{t + H}, y_{t + H}, z_{t + H})$, and $(x_t, y_t, z_t)$, which is then divided by a constant. In other words, $h_t = (x_{t + H} - x_t, y_{t + H} - y_{t}, z_{t + H} - z_t) / h_{\text{const}}$. If $t + H$ is greater than the total time stamp $T$ of the path, we simply use $T$ in place of $t + H$.
The heading vector will be part of the input to the student policy to provide guidance on the desired direction of travel. Note that we do not directly use the difference between the goal location and current location as the heading vector since the goal location can be far away from the current location and thus heavily occluded. Since our student policy does not have information about the entire terrain, it is difficult for it to learn to navigate to locations that are beyond the cameras' field of view.
The label of the path plan (i.e. the desired output from the student policy) at $s_t$ will be the next $N_T$ positions on the expert path. In other words, the label of the path plan at $s_t$ will be  $(x_t, y_t, z_t), \dots, (x_{t+N_T -1 }, y_{t+N_T -1 }, z_{t+N_T -1 })$.

In summary, each sample of training data contains the current pose of the vehicle, one RGB image and a log depth image captured at the current pose, one downward-facing RGB image captured from a tilted camera, a heading vector pointing to the desired goal location, and a path segment consisting of the next few positions generated using the expert planner.

\subsection{Student Policy Architecture}

\begin{figure}[th]
\centering
\includegraphics[trim={1cm 0cm 0cm 0cm},clip,width=8.5cm]{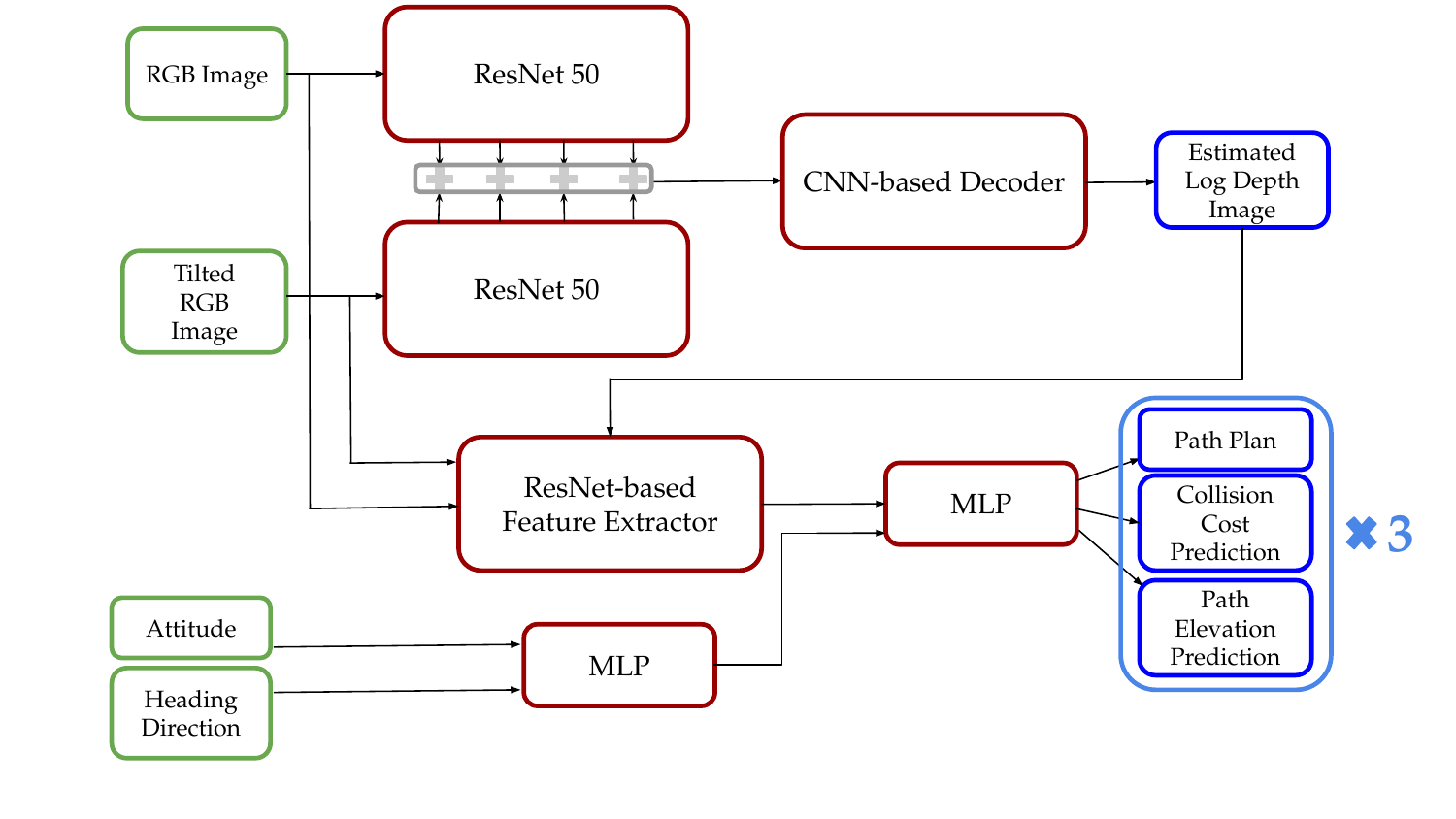}
\caption{Visualization of our policy architecture. Green boxes denote inputs, red boxes denote trainable modules, and blue boxes denote outputs from the policy. The policy takes as inputs two RGB images, the attitude of the vehicle, and a heading direction. The RGB images are used for depth estimation and also as parts of the inputs to the feature extractor. Note that instead of using the output from the final layer of ResNet, we aggregate the features from intermediate layers to capture both global and local information of the images for depth prediction. Visualization of the feature aggregation is shown in Fig \ref{fig:resnet50}.
Note that the policy produces an estimated log depth image as well as three sets of path plans together with the predicted collision cost and elevations of each planned path.}
\label{fig:student_arch}
\end{figure}

Our policy architecture is visualized in Fig \ref{fig:student_arch}. Recall that the inputs consist of two RGB images taken at current location of the vehicle, the attitude of the vehicle, and a heading direction indicating the desired travel direction. 

\begin{figure}
\centering
\includegraphics[trim={0cm 2cm 0cm 0cm},clip,width=8.5cm]{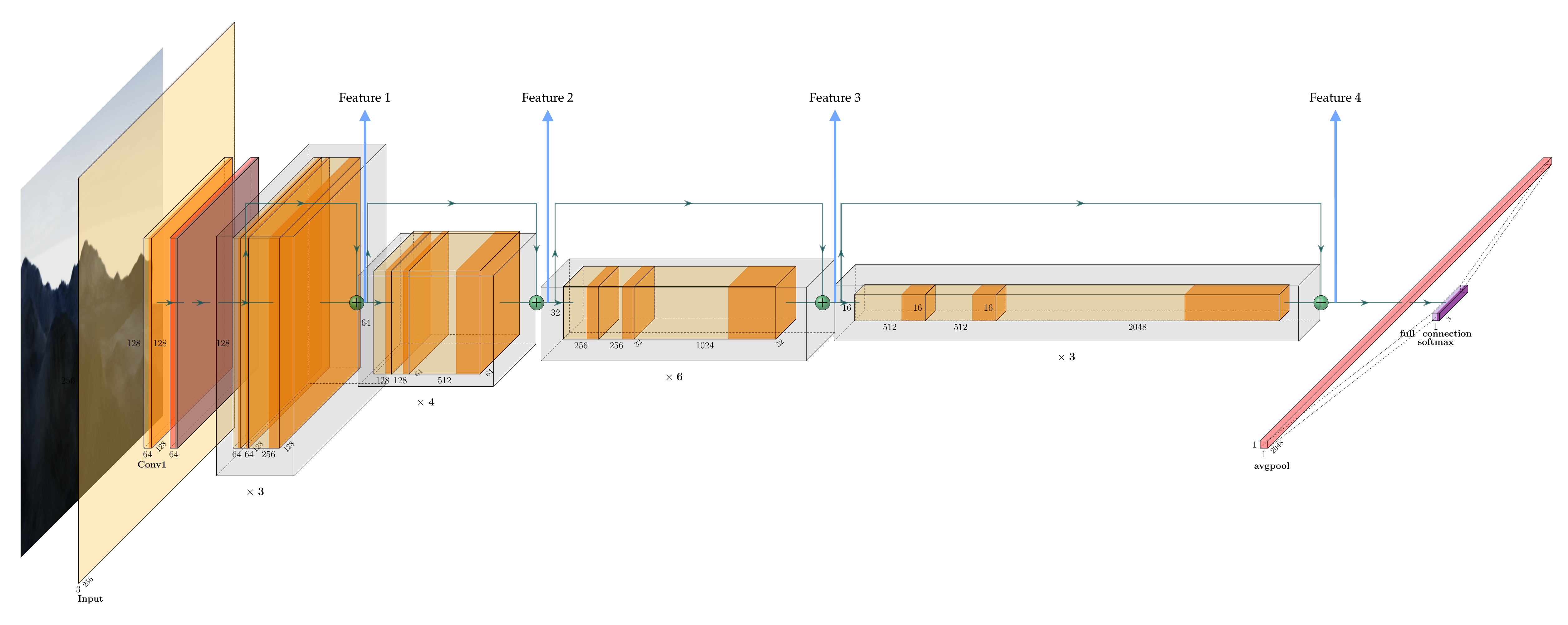}
\caption{Feature extractions from ResNet50 for depth estimation. Instead of using the output from the last layer from ResNet, inspired by \protect\cite{kendall2018multi}, we aggregate features from intermediate layers to capture both global and local information from the input image. The figure is generated using \protect\cite{harisiqbal}.}
\label{fig:resnet50}
\end{figure}

To help the policy extract distance information, we first use the two RGB images to estimate the forward log depth image. To achieve this, each RGB image is fed through a ResNet50 \cite{he2016deep} to extract visual features. Instead of using the output from the final layer of ResNet50, which is obtained through an average pooling and thus may lose local information, we fuse outputs from intermediate layers to capture both global information and local information from the image. As shown in Fig \ref{fig:resnet50}, we aggregate four features produced by four major parts of ResNet50, inspired by \cite{lin2017feature}. After we obtain four features from both the forward facing and tilted RGB images, we concatenate each pair (i.e. we obtain four concatenated features with twice the sizes) and pass each concatenated pair through a 2D convolutional layer to reduce the dimension. We call the four features resulting from this operation fused features 1, 2, 3, 4. The 4 fused features are then given to a CNN-based decoder, which consists of five blocks of 2D transposed convolutions followed by a ReLu activation, a 2D convolution, and another ReLu. The fused feature from the deepest layer of ResNet50 (i.e. fused feature 4) is passed to the first decoder block, and then the output is concatenated with the fused feature 3 and passed to the second decoder block, and similarly for the rest. The output of the decoder is the estimated log depth image of the forward-facing camera.

Next, the estimated log depth image, together with the two RGB images, is fed to a ResNet-based feature extractor to extract useful features for path planning. In practice, we find that feeding the two RGB images again to the feature extractor helps the training process compared to only feeding the estimated log depth image. The ResNet-based feature extractor takes a 7-channel input instead of 3 since we concatenate the two RGB images and the log depth image channel-wise. We removed the last fully connected layer of ResNet50.

The extracted feature is then concatenated with an embedding of the vehicle attitude and heading direction, which is obtained by passing them through a multilayer perceptron (MLP). The resulted concatenated tensor is then passed through another MLP to produce the final result, which consists of three sets of the following: a planned path, the predicted collision cost of the planned path, and the predicted elevation with respect to the terrain of each point in the path. Making the policy predict collision costs and elevations relative to the terrain helps it learn features useful for collision prediction and elevation prediction, thus helping it become more terrain-aware. Generations of ground truth labels for collision cost and elevations will be discussed in the next subsection.
We predict three paths together with collision and elevation information in order to capture the multi-modality of expert demonstration and to prevent mode collapse during training, as done in \cite{loquercio2021learning}.

\subsection{Training Objective Design}
Our training objective is a weighted sum of several terms:
\begin{align*}
{\mathcal{L}} = &k_\text{bc} {\mathcal{L}}_{\text{bc}} + k_\text{c} {\mathcal{L}}_{\text{c}} + k_{\text{alt}} {\mathcal{L}}_{\text{alt}} + k_{\text{depth}}{\mathcal{L}}_{\text{depth}} \\
&+ k_{\text{terrain}}  {\mathcal{L}}_{\text{terrain}} + k_{\text{elevation}}{\mathcal{L}}_{\text{elevation}}    
\end{align*}
where ${\mathcal{L}}_{\text{bc}}$ supervises the planar part of the planned paths (i.e. $x, y$ coordinates), ${\mathcal{L}}_{\text{c}}$ penalizes the difference between the predicted collision cost and the ground truth collision cost, ${\mathcal{L}}_{\text{alt}}$ penalizes the difference between the predicted path elevations and the ground truth path elevations, ${\mathcal{L}}_{\text{depth}}$ penalizes the difference between the predicted log depth image and the ground truth log depth image, ${\mathcal{L}}_{\text{terrain}}$ penalizes collision into the terrain, and ${\mathcal{L}}_{\text{elevation}}$ penalizes the policy for not maintaining low elevations. $k$'s are hyperparameters used to balance the terms.

{$\mathbf{{\mathcal{L}}_{\text{c}}}$:} The ground truth collision cost is computed using the method from \cite{loquercio2021learning}, which assigns collision cost to a position based on its distance to nearby points in the terrain point cloud. And the loss ${\mathcal{L}}_{\text{c}}$ is computed using mean squared error (MSE) between the ground truth collision cost and the predicted collision cost from the policy.

{$\mathbf{{\mathcal{L}}_{\text{alt}}}$:} The ground truth elevation is given by $\texttt{Elevation}_{{\mathcal{T}}}(\cdot, \cdot)$. The loss ${\mathcal{L}}_{\text{alt}}$ is computed using MSE between the ground truth elevations of the planned paths and the predicted path elevations from the policy.

{$\mathbf{{\mathcal{L}}_{\text{depth}}}$:} The loss ${\mathcal{L}}_{\text{depth}}$ is computed using MSE between the ground truth log depth and the predicted log depth from the policy.

{$\mathbf{{\mathcal{L}}_{\text{elevation}}}$:}
As explained in the previous section, to refine the elevation of predicted paths, we supervise the predicted $z$ coordinates via a self-supervision loss term:
\begin{align}
    {\mathcal{L}}_{\text{elevation}} (x, y, z) = \|z - (\delta z_{\text{des}} + \texttt{Elevation}_{{\mathcal{T}}}(x, y))\|_2^2
\end{align}

{$\mathbf{{\mathcal{L}}_{\text{bc}}}$:}  The planar motion is supervised by the relaxed winner-takes-all loss used in \cite{loquercio2021learning} to capture the multi-modality in expert demonstrations, which essentially uses the planned path that is closest to the expert to compute the loss but also adds some small weights to the planned paths to ensure differentiability. Note that only $x, y$ coordinates are used in our case instead of $x, y, z$ as in \cite{loquercio2021learning}.

{$\mathbf{{\mathcal{L}}_{\text{terrain}}}$:} To prevent the policy from producing points that are below the terrain, the loss term  $L_{\text{terrain}}$ is computed via:
\begin{align}
    {\mathcal{L}}_{\text{terrain}}(x, y, z) = \texttt{ReLu}(\texttt{Elevation}_{{\mathcal{T}}}(x, y) - z)
\end{align}
which will only become active when the point goes below the terrain.

\begin{figure}[t]
\centering
\includegraphics[trim={6cm 0cm 6cm 0cm},clip,width=8.5cm]{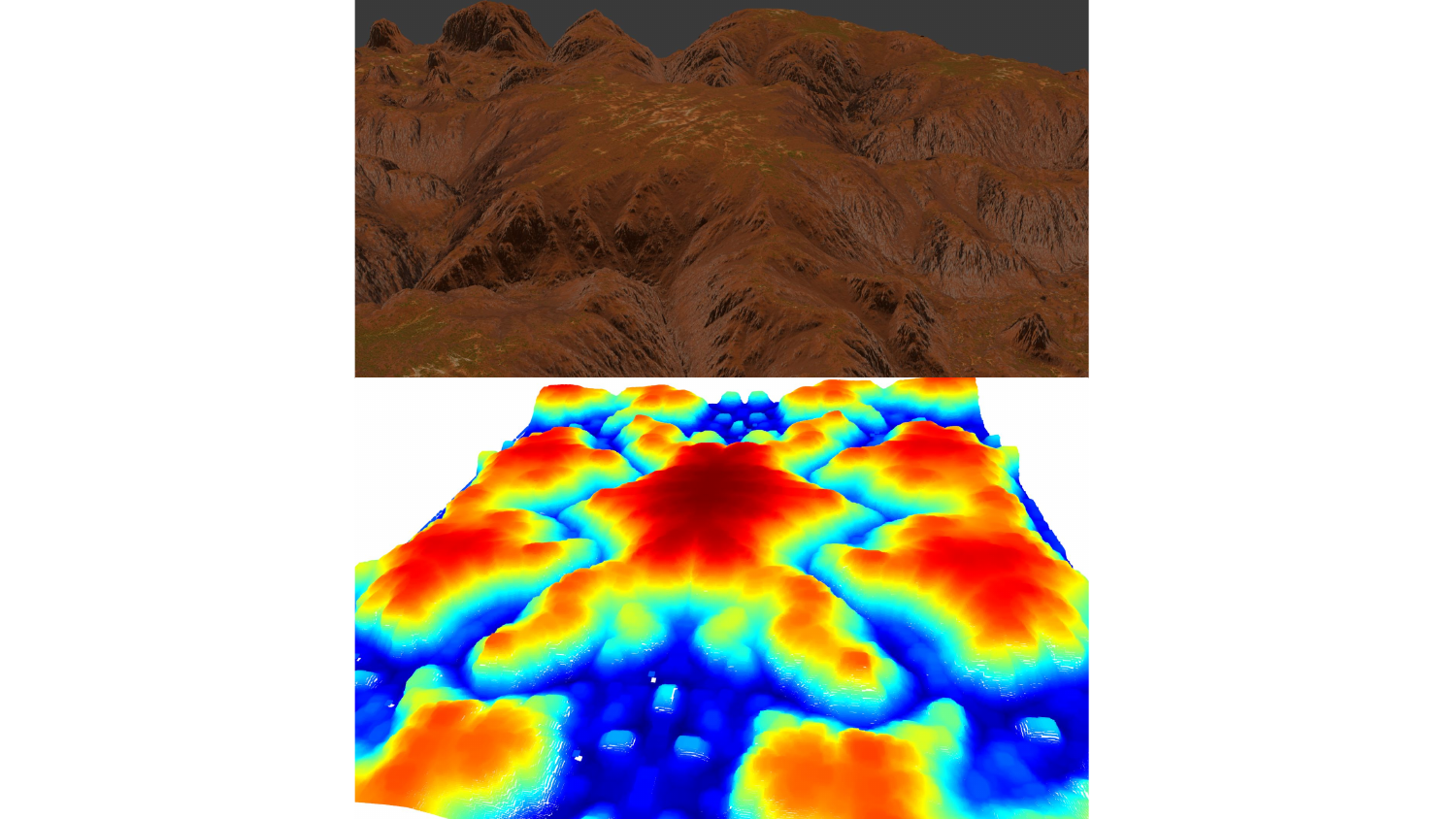}\\
\caption{{Terrain used for simulation experiments.}}
\label{fig:terrain_pcd}
\end{figure}

\section{Simulation Experiments}

In this section, we explain the setup of the simulation experiments and present the results.

\subsection{Terrain}
We build a custom terrain using Blender \cite{blender}. The terrain with the corresponding point cloud are visualized in Fig \ref{fig:terrain_pcd}, 
The terrain is exported to NVIDIA Isaac Sim to perform simulation experiments. The size of the terrain is about 820m $\times$ 820m $\times$ 60m, which is scaled down from its original size to accelerate the rendering and data collection process.

\subsection{Expert Planner Parameters}
We set $\alpha = 1.0, \beta = 0.1$ in Eq \ref{eq:expert_cost}. The minimum elevation with respect to the terrain is set to $5.0$m and the maximum elevation is set to $15.0$m. Since our terrain is scaled down, we accordingly scale down the minimum turning radius to $3$m and set the maximum climbing rate to $0.6$ rad/s. The cruise speed is set to $5$m/s. The range for RRT* is decreased to $300$m. The resolution used to check state validity is increased to $0.01$.

\subsection{Dataset Preparation}
We collect expert demonstrations on 1/4 of the terrain and use the full terrain for testing. We collect 1200 trajectories, randomly sampling start and goal locations  such that each pair of start and goal locations are separated by at least 50m. 

The parameters used for interpolating the expert trajectories are $d = 10.0$, $H = 750$, and $N_T = 10$. Note that if a point on the expert path is less than $H$ steps away from the last point on the path, we simply use the last point to compute the heading vector. Moreover, if a point on the path is less than $N_T$ steps away from the last point, we repeat the last point so that we still obtain a desired path with $N_T$ points.

To keep the inputs at similar scale, the heading vectors are divided by $h_{\text{const}} = 100.0$, the log depth images are divided by $10.0$ and RGB images are divided by $255$.

\subsection{Training}
The coefficients of the different loss terms are $k_{\text{bc}} = 0.5$, $k_{\text{c}} = 2.0$, $k_{\text{alt}} = 1.0$, $k_{\text{depth}} = {10}^6$, $k_{\text{terrain}} = {10}^3$, $k_{\text{elevation}} = 1.0$. Note that depth prediction is given a higher weight to encourage learning depth prediction first. In practice, one could also train the depth prediction module first until convergence and then freeze the depth prediction module while training the other modules. The terrain collision loss is given a high value to prevent the policy from generating points that are below the terrain. 

We use Adam optimizer \cite{kingma2014adam} with an initial learning rate $10^{-4}$, a linear decay schedule, and weight decay $10^{-5}$. The batch size is set to 32. We use $80\%$ of the data for training and the rest for validation. We train the policy until convergence, which takes about 130 epochs.

\subsection{Baseline Policy}
In addition, we also train a baseline policy for comparison study, which uses the standard behavior cloning (BC) approach. In other words, the baseline policy does not have the loss terms ${\mathcal{L}}_{\text{terrain}}$ and ${\mathcal{L}}_{\text{elevation}}$, and the loss term ${\mathcal{L}}_{\text{bc}}$ applies to all three coordinates $x, y, z$ instead of only $x, y$. The other setups are the same (i.e. dataset, architecture, and hyperparameters).

\begin{figure}
\centering
\includegraphics[trim={5cm 10cm 3cm 5cm},clip,width=8.5cm]{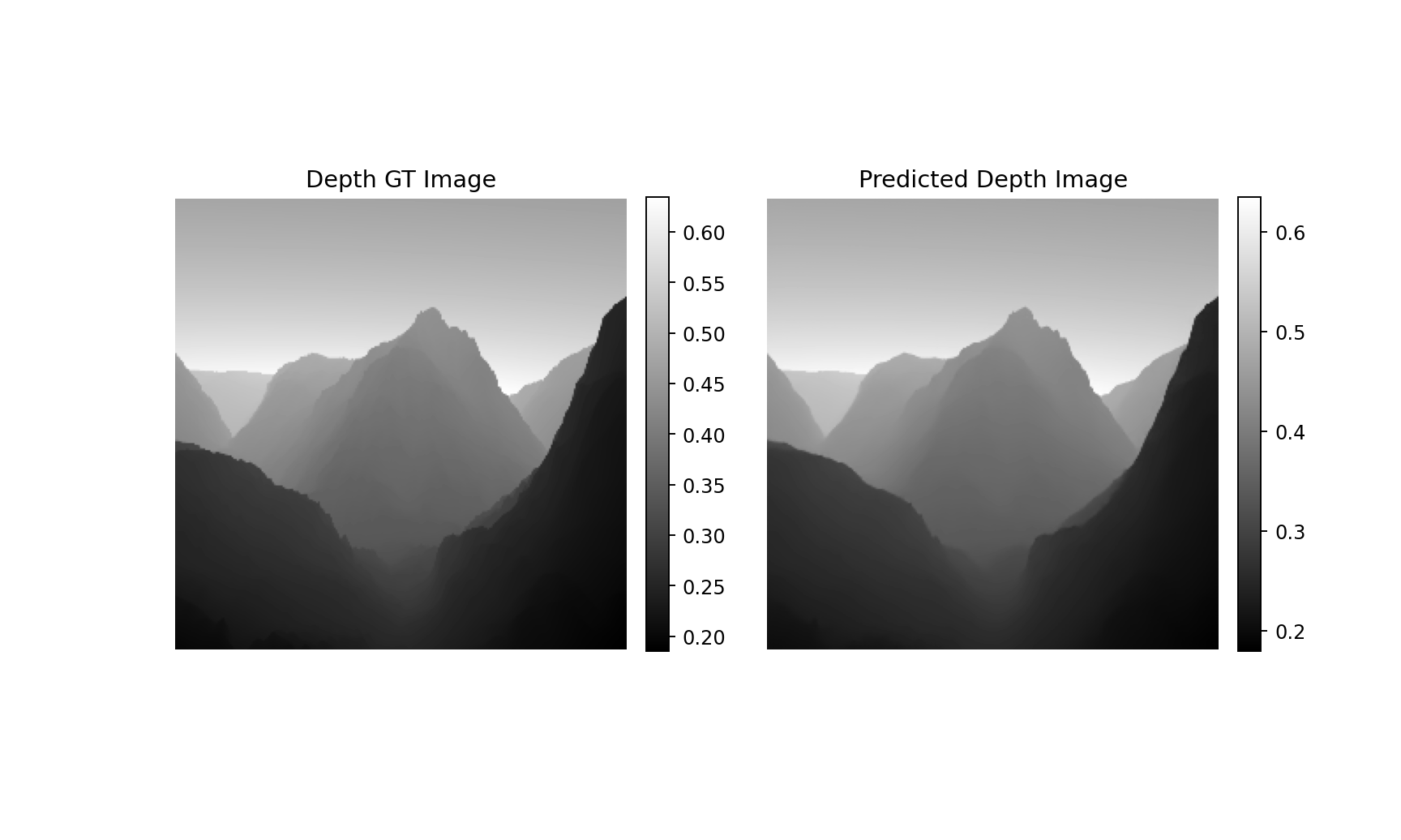}\\
\caption{{Example of ground truth (GT) log depth image versus estimated log depth image generated by the policy (both are divided by 10 to normalize the scale). The mean absolute error is about 0.0017 and the maximum absolute error (over all pixels) is 0.1622. Note that the maximum error occurs around the edges.}}
\label{fig:depth-est}
\end{figure}

\subsection{Results}
\begin{table}
\caption{\bf Quantitative Results of Simulation Experiments}
\label{tab:sim_results}
\centering
\begin{tabular}{|c c c|}
\hline
 & \bfseries BC & \bfseries Ours   \\
\hline
Average Path Length (m)  & 51.701 & 50.812 \\
Average Path Elevation (m) &  11.410  & 8.592\\
\hline
\end{tabular}
\end{table}


To compare with the baseline policy, which we term as BC, we select 20 pairs of start and goal locations that are separated by no more than 100m (since the policies were trained on heading vectors with norms less than the path length equivalent to $H$ steps, which is around $112.5$m). The start and goal locations are selected only in the unseen part of the terrain.
Since both BC and our policy output 3 modes of paths, for both policies, we execute the path with the lowest predicted collision cost.
The yaw angles of the camera are computed so that the camera always faces the goal location, and the roll and pitch angles are set to 0.

The results are recorded in Table \ref{tab:sim_results}. Note that the path elevation we report is the planned $z$-coordinates of the path rather than the relative elevation with respect to the terrain. As a reference, the average $z$-coordinate of the terrain is $18.41$m.
We observe that our policy is able to generate path plans with lower altitude than BC, while having similar path lengths.
The average depth prediction error, after converting from log scale to the original scale (in meters), is 1.82m. The average inference time is $0.0123$s with a standard deviation of $0.0009$ on an Nvidia GeForce RTX 4090 GPU.
Note that out of the 20 pairs of start and goal locations, our method successfully produced collision-free paths for all but one pair of start and goal locations, where the view at the start location is highly occluded by the mountain in front of the camera. 

Visualization of a predicted log depth image versus the ground truth log depth image is shown in Fig \ref{fig:depth-est}. We observe that the predicted depth images are generally good at capturing the distance information of large objects (e.g. main bodies of the mountains) but may fail to accurately predict detailed information such as distances to trees on the mountain.

Visualization of a comparison between the path generated by the expert and the path generated by our policy is shown in Fig \ref{fig:expert_vs_student}. Note that our policy is able to generate paths lower than the expert demonstrations, since we supervised the elevation of the path plans directly using the terrain.




\section{Conclusions}
In this paper, we proposed a method to train a policy that is capable of low-altitude path planning with only RGB images and attitude information. Our method combines BC and self-supervised learning to enable the learned policy to outperform the student policy trained with only a standard BC approach.


One limitation of the current method is that it does not explicitly take into account the max climb rate of the aircraft during training. Potential solutions include adding a differentiable optimization-based tracking controller (e.g. differentiable MPC) after the student policy and training the composed modules end-to-end as done in \cite{li2024ikap}.


\bibliographystyle{IEEEtran} 
\bibliography{main}






\end{document}